\documentclass[journal,twoside,web]{ieeecolor}
\usepackage{generic}
\usepackage{cite}
\usepackage{fancyhdr}
\usepackage{amsmath,amssymb,amsfonts}
\usepackage{algorithmic}
\usepackage{graphicx}
\usepackage{algorithm,algorithmic}
\usepackage{hyperref}
\hypersetup{hidelinks=true}
\usepackage{textcomp}
\usepackage{xcolor}
\usepackage{threeparttable}
\usepackage{multirow}
\usepackage{booktabs}
\usepackage{upgreek}

\def\BibTeX{{\rm B\kern-.05em{\sc i\kern-.025em b}\kern-.08em
    T\kern-.1667em\lower.7ex\hbox{E}\kern-.125emX}}
\begin{document}
\title{Eat-Radar: Continuous Fine-Grained Intake Gesture Detection Using FMCW Radar and 3D Temporal Convolutional Network with Attention}
\author{
Chunzhuo Wang, \IEEEmembership{Student Member, IEEE}, T. Sunil Kumar, \IEEEmembership{Member, IEEE}, Walter De Raedt,\\ Guido Camps, Hans Hallez, \IEEEmembership{Member, IEEE}, Bart Vanrumste, \IEEEmembership{Senior Member, IEEE} 
\thanks{This work is funded in part by KU Leuven under grant C3/20/016, in part by Goodbrother project (Network on Privacy-Aware Audio- and Video-Based Applications for Active and Assisted Living with COST Action 19121), and in part by the China Scholarship Council (CSC) under grant 202007650018. \emph{(Corresponding author: Chunzhuo Wang.)}}
\thanks{Chunzhuo Wang, and Bart Vanrumste are with the e-Media Research Lab, and also with the ESAT-STADIUS Division, KU Leuven, 3000 Leuven, Belgium (e-mail: chunzhuo.wang@kuleuven.be; bart.vanrumste@kuleuven.be).}
\thanks{T. Sunil Kumar is with Vellore Institute of Technology, Chennai, India (e-mail: telagamsetti.sunilkumar@vit.ac.in).}
\thanks{Walter De Raedt and Chunzhuo Wang are with the Life Science Department, IMEC, 3001 Heverlee, Belgium (e-mail: Walter.DeRaedt@imec.be).}
\thanks{Guido Camps is with the Division of Human Nutrition and Health, Department of Agrotechnology and Food Sciences, Wageningen University and Research, 6700EA Wageningen, and also with the OnePlanet Research Center, 6708WE Wageningen, The Netherlands (e-mail: guido.camps@wur.nl).}
\thanks{Hans Hallez is with the M-Group, DistriNet, Department of Computer Science, KU Leuven, 8200 Sint-Michiels, Belgium (e-mail: hans.hallez@kuleuven.be).}
}
\maketitle
\thispagestyle{fancy}

\begin{abstract} 
Unhealthy dietary habits are considered as the primary cause of various chronic diseases, including obesity and diabetes. The automatic food intake monitoring system has the potential to improve the quality of life (QoL) of people with diet-related diseases through dietary assessment. In this work, we propose a novel contactless radar-based approach for food intake monitoring. Specifically, a Frequency Modulated Continuous Wave (FMCW) radar sensor is employed to recognize fine-grained eating and drinking gestures. The fine-grained eating/drinking gesture contains a series of movements from raising the hand to the mouth until putting away the hand from the mouth. A 3D temporal convolutional network with self-attention (3D-TCN-Att) is developed to detect and segment eating and drinking gestures in meal sessions by processing the Range-Doppler Cube (RD Cube). Unlike previous radar-based research, this work collects data in continuous meal sessions (more realistic scenarios). We create a public dataset comprising 70 meal sessions (4,132 eating gestures and 893 drinking gestures) from 70 participants with a total duration of 1,155 minutes. Four eating styles (fork \& knife, chopsticks, spoon, hand) are included in this dataset. To validate the performance of the proposed approach, seven-fold cross-validation method is applied. The 3D-TCN-Att model achieves a segmental F1-score of 0.896 and 0.868 for eating and drinking gestures, respectively. The results of the proposed approach indicate the feasibility of using radar for fine-grained eating and drinking gesture detection and segmentation in meal sessions.

\end{abstract}


\section{Introduction}
\label{sec:introduction}

\IEEEPARstart{U}{nhealthy} dietary habits are closely related to multiple chronic health problems such as overweight, diabetes, malnutrition, eating disorder, and cardiovascular diseases \cite{b1}. These diet-related diseases can not only affect individuals' daily life but also result in substantial financial and societal costs \cite{b2}. According to WHO reports, the overweight and obesity have been considered as a global public health problem affecting people's quality of life (QoL)\cite{b3}, and imbalanced diet is the primary cause of this issue. In order to keep a healthy diet, it is necessary to track dietary intake.

Recently, different automatic food intake monitoring systems have been proposed for this purpose. These solutions can be categorized into 3 types: video-based, wearable sensor-based, and portable sensor-based approaches. The video-based method \cite{b4, b23, b24} uses cameras to detect eating gestures. The wearable sensor-based approach mainly utilizes an inertial sensor \cite{b5,b18,b40}, a miniature camera \cite{b25}, an acoustic sensor \cite{b26} or integrates multiple sensors \cite{b41} to monitor hand-to-mouth movements or chewing movements. The smart plate \cite{b6}, the dining tray \cite{b32}, and the smart table surface \cite{b27} are categorized into the portable sensor-based approach. Existing methods have been proven feasible for food intake monitoring; however, some disadvantages can emerge. The wearable sensors require precise placement, which can be uncomfortable for long-term duration. Battery concerns are also associated with such sensor types. As for video-based methods, issues related to lighting conditions and privacy may arise. In order to account for these drawbacks, we propose a contactless, privacy-preserving radar-based approach for in-home food intake monitoring.

Recently, radar, as an emerging technology, combined with machine learning, has been gradually applied to a variety of scenarios, such as health monitoring \cite{b28, b29}, and human activity recognition (HAR) \cite{b39}. Much radar-based HAR research in indoor/kitchen scenarios has been conducted \cite{b7, b8, b9}. Luo \textit{et al.} \cite{b7} used the Doppler-Time map (DT map) from two radars and a convolutional neural network (CNN) to recognize 15 activities (eating activity included) from 3 participants in a kitchen environment, with an overall classification rate of 92.8\%. Gorij \textit{et al.} \cite{b8} deployed two radars to classify 8 activities from 12 participants. They manually extracted features and used a Random Forest (RF) classifier, with a weighted F1-score of 93.63\%. Maitre \textit{et al.} \cite{b9} utilized Range-Time maps (RT map) from three ultra-wideband (UWB) radars and applied a CNN and long-short-term-memory network combined model (CNN-LSTM) to recognize 15 different activities from 10 participants, yielding an F1-score of 93\% for eating activity detection. It should be noted that the above research mainly focused on multi-activities recognition. For each activity, they can classify the session/period coarsely. Specifically, they recognized the eating session from the start to the end, but did not recognize fine-grained eating gestures. 

The fine-grained eating gesture recognition includes details such as how many eating gestures happened in that period (counting the number of eating/drinking gestures) and the duration of each detected gesture. In this study, the eating/drinking gesture is defined as the movement from raising the left/right hand to the mouth with the fork/knife/spoon/chopsticks/containers until putting away the hand from the mouth.

Xie \textit{et al.} \cite{b10} were the first to explore the potential of using mmWave Frequency Modulated Continuous Wave (FMCW) radar sensors for fine-grained eating behavior monitoring. They collected eating gestures with five classes (fork, fork \& knife, spoon, chopstick, and bare hand) from 6 participants. They detected eating gestures by searching the repetitive pattern of hand and arm movements that bring food from the table to mouth. Once an eating gesture was detected, an unsupervised clustering method was utilized to segment the detected eating gesture. After segmentation, a CNN-based classifier was used to classify the segmented data. In a recent work \cite{b33}, the authors collected a total of 324 eating and drinking gestures from 6 participants in a standing scenario. Using CNN-Bidirectional LSTM (CNN-BiLSTM) model, they achieved accuracies of 84\% and 76\% for the detection of eating and drinking activities, respectively, under Leave-One-Subject-Out (LOSO) validation.

A more complex challenge arises when dealing with real-life continuous meal sessions. Previous works on radar-based eating activity recognition collected data in limited conditions with short durations. In \cite{b7,b8,b9}, each participant's eating and drinking duration was less than 5 min. However, the real meal session has a longer duration (10-20 min per meal in our case, Table \ref{meal_data}). Additionally, meal sessions naturally involve lots of non-feeding activities seamlessly intertwined with intake gestures \cite{b30}, resulting in a more realistic and challenging scenario for radar-based intake gesture detection. In this paper, we build a new end-to-end approach for fine-grained intake gesture detection using a contactless mmWave FMCW radar and a sequence-to-sequence (seq2seq) 3D temporal convolutional network with self-attention (3D-TCN-Att). The key highlights of this work are as follows:

\begin{enumerate}

\item A radar-based system is applied in the food intake monitoring domain to detect eating gestures in meal sessions. A seq2seq 3D-TCN-Att model is applied to process the Range-Doppler Cube (RD Cube) and generate frame-wise predictions. Combined with IoU-based segment-wise evaluation, this method is able to not only count the number of intake gestures, but also segment the time intervals of intake gestures.
\item We complete an intensive comparison between the proposed approach to the state-of-the-art techniques reported in the literature. Firstly, different radar data representations, namely RD Cube and DT map, are selected to find the most suitable one based on machine learning performance. Secondly, different models: CNN-(Bi)LSTM, pretrained 3D-CNNs, and transformers, are used to compare the performance. Results demonstrate the feasibility of radar-based intake gesture detection.
\item We produce a publicly available dataset\footnote{https://rdr.kuleuven.be/dataset.xhtml?persistentId=doi:10.48804/NQSH4F} consisting 70 meal sessions from 70 participants. In total, 4,132 eating gestures and 893 drinking gestures are recorded. To our knowledge, this is the first public radar dataset targeting eating gesture detection in real meal sessions.
\end{enumerate}

\section{Methods}
\subsection{FMCW Radar Principle}
For an FMCW radar with single transmitter (TX) and single receiver (RX), as called single-input-single-output (SISO) radar, its TX periodically transmits a frequency modulated sinusoidal signal with bandwidth $B$, start frequency $f_c$, frequency slope $S$, period $T_c$, as shown in Fig. \ref{radar-theory}. The signal in one periodic time is called a chirp signal. A sequence of chirps represents a frame (Fig. \ref{radar-theory}(c)(d)). The row dimension of the frame is called the fast time dimension, and the column dimension is known as the slow time dimension, as the time scale between two chirps is much larger than the time scale between two sample points within a chirp. The RX antenna receives the signals reflected from objects in the radar's field of view (FOV). By mixing the received signal and transmitted signal, the mixer generates the Intermediate Frequency (IF) signal, also called the beat signal. The instantaneous frequency ($f_{b}$) of IF signal is equal to the difference of the instantaneous frequency between the transmitted and received signal. Similarly, the phase of IF signal ($\phi$) is the difference of phase between the transmitted and received signal.
\subsubsection{Range Measurement}
{
For a static object with a distance $d$ to the radar (Fig. \ref{radar-theory}(b)), the time delay of the received signal is $\tau=\frac{f_{b}}{S}=\frac{2d}{c}$, where $c$ is the speed of light. Hence, the relation between the range $d$ and beat frequency $f_{b}$ is $d=\frac{f_{b}c}{2S}$. To calculate the range, a Fast Fourier Transform (Range-FFT) is applied to convert the IF signal from the time domain to the frequency domain for each chirp. The initial phase ($\phi$) of the IF signal is $\phi=2\pi f_c\tau=\frac{4 \pi d}{\lambda}$, where $\lambda$ is the wavelength.
}

\subsubsection{Velocity Measurement}
\begin{figure}[t]
      \centering
      \includegraphics[scale=0.48]{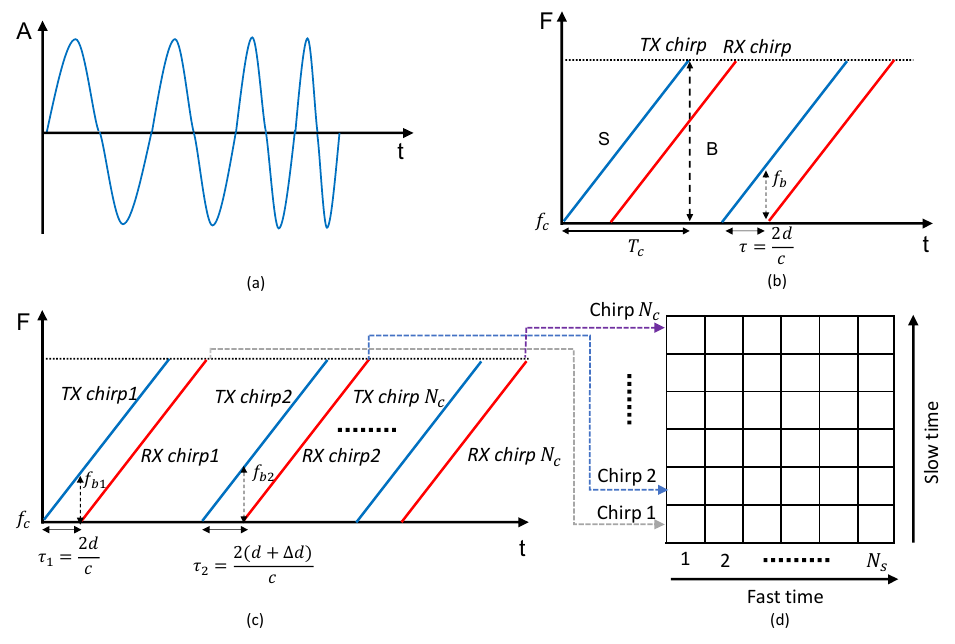}  
      \vspace{-0.3cm}    
      \caption{(a) An example of the time-amplitude response from radar TX signal. The x-axis represents the time (t), and the y-axis represents the amplitude (A) of the signal. (b) The time-frequency response of TX signal (blue line) and RX signal (red line). The x-axis represents the time (t), and the y-axis represents the frequency (F). $B$, $f_c$, $S$, $T_c$ are the bandwidth, start frequency, slope, and period of TX signal. $f_{b}$, $d$, $c$, $\tau$ are the beat frequency, the distance between the radar and the object, the speed of light, and the time delay between TX and RX signal. (c) The consecutive TX and RX signals for a moving object. $\tau_1$, $\tau_2$ represent the time delay for the first and second chirp, $\Delta d$ is the displacement of the moving target in one TX period. (d) The form of the initial data frame after mixing TX and RX signals. Each row represents the $N_s$ sample points for a chirp after mixing and analog to digital conversion (ADC), $N_c$ chirps are stacked together to form the initial data frame. The row dimension of the frame is called the fast time dimension, and the column dimension is known as the slow time dimension, as the time scale between two chirps is much larger than the time scale between two sample points in a chirp.}  \vspace{-0.2cm}
      \label{radar-theory} 
\end{figure}

{
For a moving object with range $d$ and velocity $v$, there is a tiny displacement $\Delta d$ between two consecutive chirps, as shown Fig. \ref{radar-theory}(c). This causes an IF frequency change $\Delta f_b=f_{b2}-f_{b1}=S(\tau_2-\tau_1)=S(\frac{2(d+\Delta d)}{c}-\frac{2d}{c})=\frac{2S\Delta d}{c}$ and a phase change $\Delta \phi=\frac{4\pi\Delta d}{\lambda}=\frac{4\pi vT_c}{\lambda}$. Normally, the $\Delta f_b$ caused by $\Delta d$ is negligible compared to the scale of the sweep bandwidth (B), whereas the phase change is more sensitive to the object's movement between chirps in one frame. Hence, to estimate the velocity, an FFT across the chirps (along slow time axis, see Fig. \ref{radar-theory}(d)) for each range bin in one frame is applied after Range-FFT processing, as called Doppler-FFT. More details of the FMCW radar principle are in the website of Texas Instruments (TI)\footnote{See https://training.ti.com/node/1139153}.
}

It should be noted that an analog to digital conversion (ADC) is used to digitize the beat signal before any processing. Suppose the number of samples in a chirp is $N_s$ after the ADC and the number of chirps in a frame is $N_c$, the dimension of a radar data frame $R$ is $N_c\times{N_s}$ (Fig. \ref{radar-theory}(d)).

\subsection{Radar System Information}

\begin{figure}[t]
      \centering
      \includegraphics[scale=0.38]{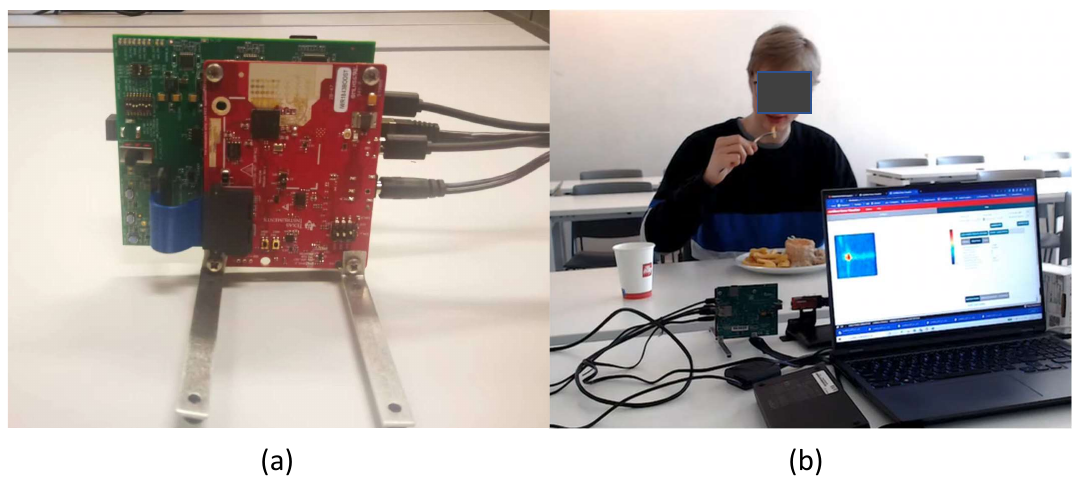}
      \vspace{-0.2cm}
            \caption{(a) The TI IWR1843 radar board (Red) and DCA1000EVM board (Green). (b) The scene of data collection. The participant was eating with a pair of fork \& knife. The radar was placed on the table.}  \vspace{-0.1cm}
      \label{experiment-scene} 
\end{figure}

The experiment utilizes a 76 GHz TI IWR1843 radar board and a DCA1000EVM data capture card to build the data collection system, as shown in Fig. \ref{experiment-scene}. The configurations of the radar system are specified in Table \ref{radar_config}. The radar has a 4 GHz bandwidth, with 3 TX and 4 RX antennas, powered with 5 V. The power consumption under described configuration is 1.8 W. The system generates 25 frames per second (fps). Two USB interfaces (one for IWR1843 radar board, one for DCA1000EVM) are connected to the laptop for system configuration. One Ethernet cable with high-speed data transmission capability (1 Gbps) is used to transfer raw data from the radar to laptop via the software TI mmWave Studio\footnote{See https://www.ti.com/tool/MMWAVE-STUDIO}.

\begin{table}[t]

\caption{Configurations of FMCW radar}
\vspace{-0.2cm}
\label{radar_config}
\begin{center}
\scalebox{0.78}{
\begin{tabular}{lr}
\toprule
 \hline
Parameter & Configuration\\
\midrule
Start frequency & 77 GHz \\
Sweep bandwidth & 3.8 GHz \\
Sweep slope  & 50 MHz/$\upmu$s\\
Frame rate  & 25 fps\\
Sample rate & 2,000 ksps \\
 Number of chirps in one frame & 128\\
 Number of samples in one chirp & 128\\
 Number of used transmitter, receivers & 1, 4\\
 Range resolution & 4 cm\\
 Velocity resolution & 4 cm/s\\
  \hline
\bottomrule
\end{tabular}}
\end{center}
\vspace{-0.3cm}
\end{table} 

\begin{figure}[t]
      \centering
      \includegraphics[scale=0.32]{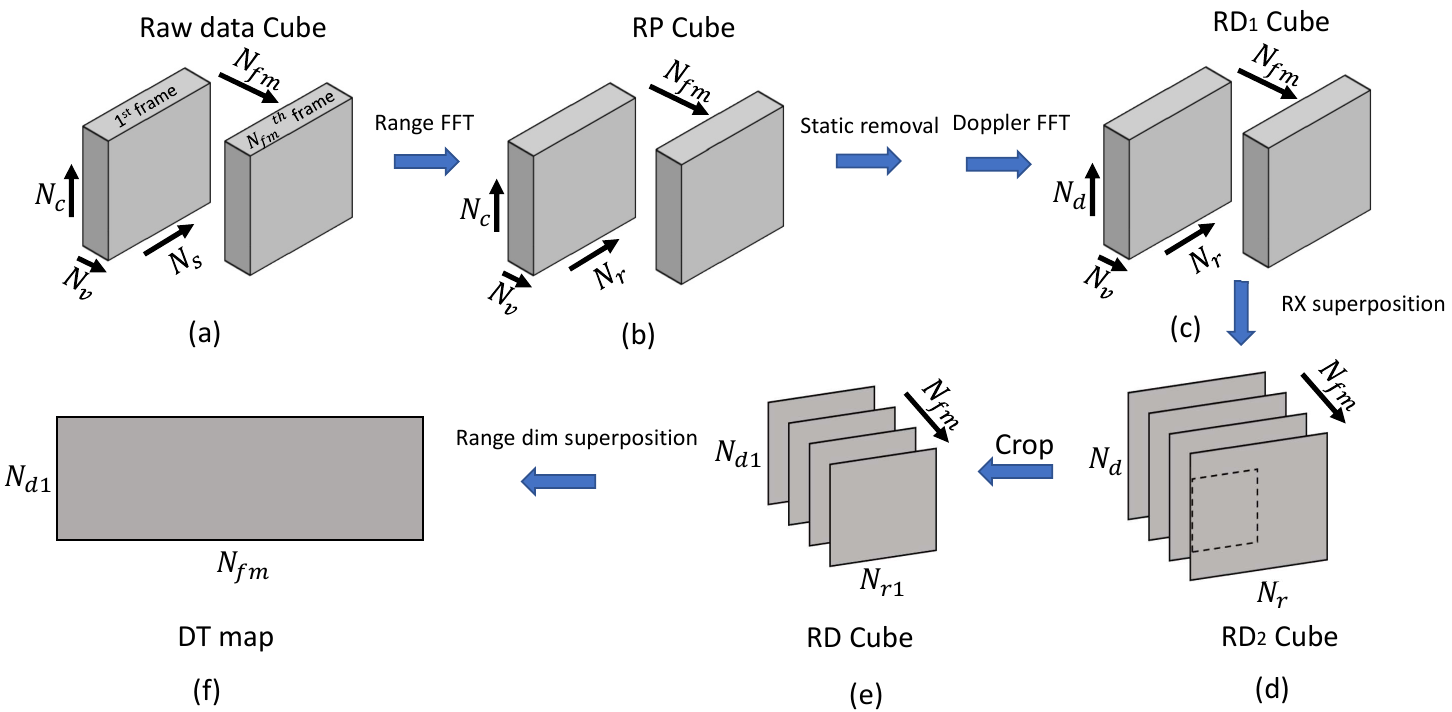}
      \vspace{-0.2cm}
      \caption{The radar signal pre-processing pipeline. (a) The raw data cube. (b) The range profile cube (RP Cube). (c) The RD$_1$ Cube. (d) The RD$_2$ Cube after RX superposition. (e) The cropped RD Cube (to reduce the computation cost). (f) The Doppler-Time map (DT map).}   \vspace{-0.2cm}
      \label{radar_process}
\end{figure} 

\begin{figure}[t]
      \centering
      \includegraphics[scale=0.33]{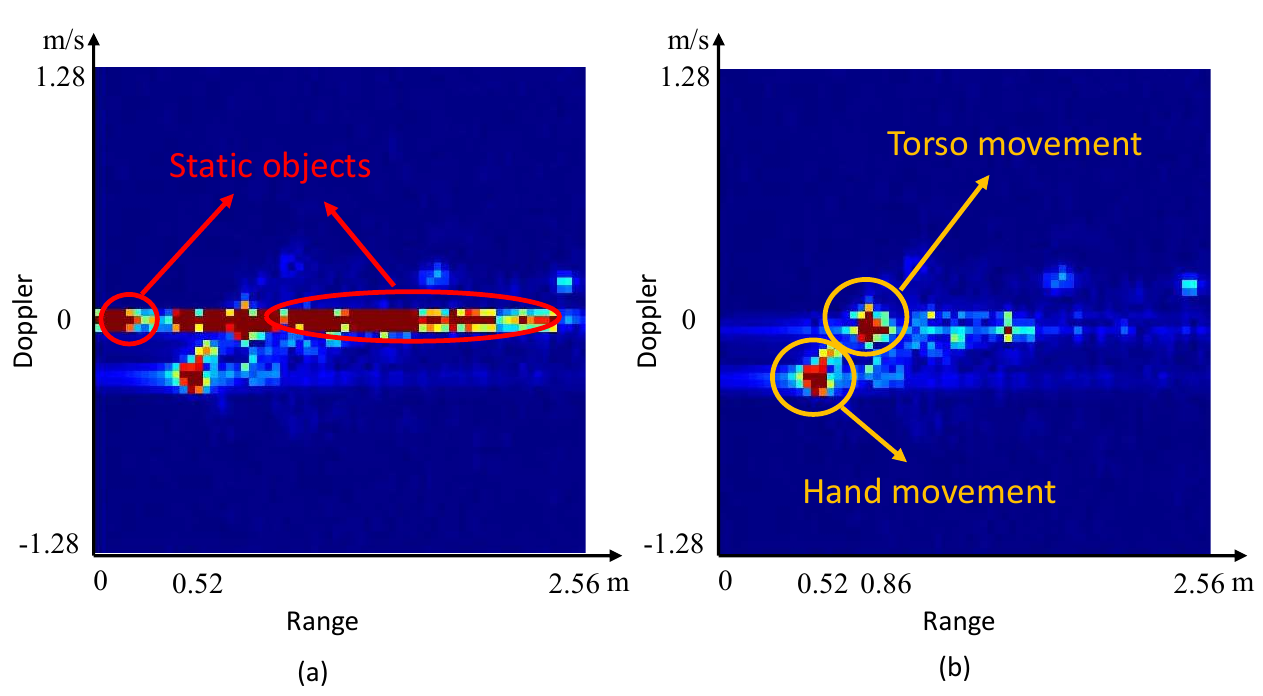}
      \vspace{-0.2cm}
      \caption{The range-doppler (RD) frame represents the transient movement taking a glass of water from table to mouth. (a) the RD frame without static object removal; (b) the RD frame after static object removal. This is a movement from plate to mouth (the hand is moving away from radar), so it has negative velocity. The torso (body) is moving towards the radar with lower velocity compared to the hand.}   \vspace{-0.2cm}
      \label{static_removal}
\end{figure}

\subsection{Radar Signal Processing}
The raw data exported from the radar are defined as a 5D data cube of size $N_{fm}\times{N_{tx}}\times{N_{rx}}\times{N_{c}}\times{N_{s}}$ where $N_{fm}$, $N_{tx}$, $N_{rx}$, $N_{c}$ and $N_{s}$ are the numbers of the radar frames in one meal session, transmitters (1), receivers (4), chirps in one frame (128), and sample points (128) in one chirp, respectively. Considering that we only use 1 transmitter and 4 receivers, the term $N_{tx}\times{N_{rx}}$ is replaced with $N_v$ which denotes the number of virtual antennas ($N_v=4$). Hence, a 4D data cube $N_{fm}\times{N_v}\times{N_{c}}\times{N_{s}}$ represents the raw data of a meal session. As mentioned in Section \uppercase\expandafter{\romannumeral2}-A, a frame of a SISO radar is represented by $R$ (with size $N_c\times{N_s}$, Fig. \ref{radar-theory}(d)), the 4D data cube can also be written in $N_{fm}\times{N_v}\times{R}$. Subsequently, several steps are applied to extract the meaningful features, as shown in Fig. \ref{radar_process}.
\subsubsection{Range FFT}
{
For each frame $R$, we apply a FFT transform along fast-time axis (in each chirp, see Fig. \ref{radar-theory}(d)), obtained range profile cube (RP Cube) with size $N_{fm}\times{N_v}\times{N_{c}}\times{N_{r}}=N_{fm}\times{N_v}\times{R_1}$, $N_{r}$ represents the number of range bins after range-FFT, $R_1$ represents the new frame in range profile cube (RP Cube). In this study, $N_{r}=N_{s}=128$.
}
\subsubsection{Static Objects Removal}
{
To remove the signal of static objects, for each frame $R_1$ in the RP Cube, we calculate the average value through all chirps (along the slow-time axis, see Fig. \ref{radar-theory}(d)), and every original chirp is subtracted by the mean chirp as follows \cite{b12}:
 \begin{equation}R_2(i,j)=R_1(i,j)-\frac{1}{N_c}\sum_{k=1}^{N_c}{R_1{(k,j)}}\label{eq1}\end{equation}
 where $R_1(i,j)$ represents the value at $i$-th chirp ($i \in [1,N_c]$) and $j$-th range bin ($j \in [1,N_r]$) in frame $R_1$, $R_2$ represents the new frame after removing static objects.
}
\subsubsection{Doppler FFT}
{
For each frame $R_2$, we apply a FFT along slow-time across chirps to generate a Range-Doppler cube (RD$_1$ Cube) with size $N_{fm}\times{N_v}\times{N_{d}}\times{N_{r}}=N_{fm}\times{N_v}\times{R_3}$, $N_{d}$ represents the number of doppler bins after doppler-FFT, $R_3$ represents the new frame in RD$_1$ Cube. In this study, $N_{d}=N_{c}=128$. 
}

\begin{figure*}[t]
      \centering
      \includegraphics[scale=0.87]{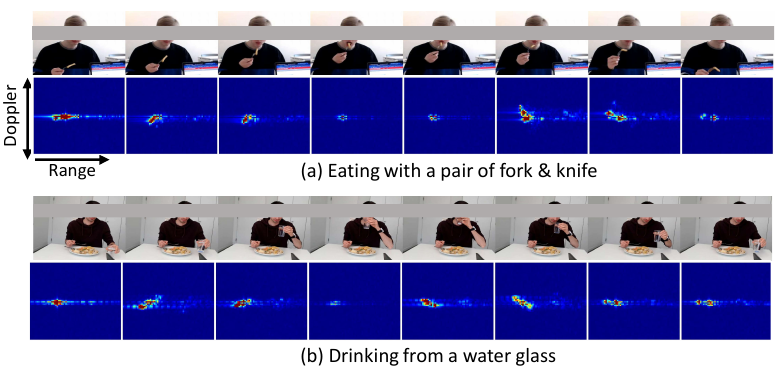}    
      \vspace{-0.2cm}
      \caption{The video frames and the corresponding RD frames of eating and drinking gestures. For the RD frames, the x-axis represents the range dimension; the y-axis represents the doppler dimension. It should be noted that the actual amounts of frames for each eating/drinking gesture are different, and we manually selected the fixed number of frames for each gesture in this figure.} \vspace{-0.2cm}
      \label{rd_scene}
\end{figure*}

\begin{table}[t]
\caption{Meal data statistics}
\vspace{-0.2cm}
\label{meal_data}
\begin{center}
\scalebox{0.8}{
\begin{tabular}{lr}
\toprule
 \hline
 Parameter & Values\\
\midrule
 Participants  &70 \\
 Fork \& knife: chopsticks:spoon:hand & 27:16:15:12\\
 Female : male & 25 : 45 \\
 Duration ratio of other : eating : drinking& 11.21:2.59:1\\
 Mean meal duration &(16.50$\pm$5.24) min \\
 Total duration & 1,155 min\\
\hline
\bottomrule
\end{tabular}}
\end{center}
\vspace{-0.3cm}
\end{table} 

\begin{table}[t]
\caption{Details of different eating styles}
\vspace{-0.2cm}
\label{eat_data}
\begin{center}
\scalebox{0.8}{
\begin{tabular}{lcc}
\toprule
\hline
Style & \#Eating gesture & Food\\
\midrule
Fork \& knife  &    1,632  &  salad, fries, chicken, beef burger, cake\\
Chopsticks     &    1,479  &  rice, fried noodles, chicken, vegetables\\
Spoon          &    671   &  rice, chicken, vegetables\\
Hand           &    350   &  chicken sandwich, egg sandwich\\
\hline
\bottomrule
\end{tabular}}
\end{center}
\vspace{-0.3cm}
\end{table} 

\begin{table}[t]
\caption{Eating and Drinking gesture Details}
\vspace{-0.2cm}
\label{eat_drink_data}
\begin{center}
\scalebox{0.77}{
\begin{tabular}{lcc}
\toprule
\hline
\multirow{2}*{Parameter} & \multicolumn{2}{c}{Values}\\
 ~& Eating & Drinking\\
\midrule
 Number of gestures& 4,132& 893\\

 Duration range of gestures & 1.00-17.08 s& 1.88-24.56 s\\
 Median duration of gestures & 2.56 s& 4.64 s\\
 Mean $\pm$ std of gestures & 2.93 $\pm$ 1.42 s& 5.24 $\pm$ 2.34 s\\
\hline
\bottomrule
\end{tabular}}
\end{center}
\vspace{-0.3cm}
\end{table} 

\subsubsection{RX Superposition}
{
Because of the narrow width (less than 1m) of the moving target (human), the angular resolution obtained by 4 receiving antennas is insufficient to differentiate objects in the cross-range dimension. Therefore, we simply superpose the RD$_1$ Cubes from 4 RX channels into one with an improved signal-to-noise ratio (SNR) \cite{b13}. The new RD$_2$ Cube has size $N_{fm}\times{N_{d}}\times{N_{r}}=N_{fm}\times{R_4}$, $R_4$ represents the new 4-in-1 RD frame:
 \begin{equation}RD_2(t,i,j)=\frac{1}{N_v}\sum_{a=1}^{N_v}{|RD_1{(t,a,i,j)}|}\label{eq2}\end{equation}
  where $RD_2(t,i,j)$ represents the value at $t$-th frame ($t \in [1,N_{fm}]$) $i$-th doppler bin ($i \in [1,N_d]$) and $j$-th range bin ($j \in [1,N_r]$), $a \in [1,N_v]$.
}
\subsubsection{Crop ROI}
{
The obtained RD$_2$ Cube contains frames with size $128\times128$. Because this study focuses on short-range HAR, and intake behavior has low speed, it is observed that the long-range area and high-velocity area are never reached. To reduce the computation cost, we crop the frames from RD$_2$ Cube to only keep the region of interest (ROI). The cropped frame size is $N_{d1}\times{N_{r1}}$, where $N_{d1}=64$ (max velocity: $\pm 1.28\ m/s$) and $N_{r1}=32$ (max range: $1.28\ m$). After this step, the final RD Cube is prepared as 3D input data with size $N_{fm}\times{N_{d1}}\times{N_{r1}}$  ($N_{fm}\times{64}\times{32}$). Fig. \ref{static_removal} shows the example of an RD frame.
}

\subsubsection{Generate the Doppler-Time Map For Comparison}
{
This extra step is used to generate the DT map for data representation comparison, which will be discussed in Section \uppercase\expandafter{\romannumeral3}-C. The DT map is generated by summing the absolute value of the signal over the range dimension \cite{b22}. 
 \begin{equation}DT(t,n)=\frac{1}{N_{r1}}\sum_{j=1}^{N_{r1}}{RD{(t,n,j)}}\label{eq3}\end{equation}
  where $DT(t,n)$ represents the DT map value at $t$-th frame ($t \in [1,N_{fm}]$) and $n$-th doppler bin ($n \in [1,N_{d1}]$), $RD{(t,n,j)}$ represents the RD Cube value at $t$-th frame, $n$-th doppler bin, and $j$-th range bin ($j \in [1,N_{r1}]$). The dimension of the DT map is $N_{fm}\times{N_{d1}}$.
}

\subsection{Data Collection and Annotation}
The ethical committee of KU Leuven has approved this research (Reference number: G-2021-4025-R4), and written informed consent from each participant was collected. 70 participants (25 females and 45 males) were recruited in the experiment. Each of them had one meal session. The data were collected in seven different rooms with different layouts at the KU Leuven University. The participants sat directly in front of the radar and ate their meals. The distance between the food plate and the radar was around 0.3 m to 0.6 m, as shown in Fig. \ref{experiment-scene}(b). A camera was used to record video for annotation. It should be noted that some of the participants also wore IMU wristbands on two hands for another study, which will not be discussed here. Table \ref{meal_data} shows detailed information of meals.

The consumed food was served by Alma (the student restaurant), a Chinese restaurant, a sandwich shop, and a convenience store. The participants could use forks \& knives, chopsticks, spoons, and hands to eat the meal. They were allowed to eat at their own pace. They could stop, talk, watch smartphones, move for a while, and prepare/stir food. A glass of water or a can of soft drink was served together with the meal. During the meal, the participants could drink at their own pace. However, to increase the number of drinking gestures, some participants were required to drink several times extra before and after the meal. This requirement is due to the fact that people have few drinking gestures compared to the amount of eating and other gestures in a meal. Detailed information on eating styles is shown in Table \ref{eat_data}.

All participants were encouraged to eat their meals in their own way. Thus some non-feeding movements were naturally included during meal sessions. Additionally, to further mimic free-living environments, 22 of 70 meal sessions were collected by asking the participants to perform more non-feeding gestures and confounding gestures \cite{b40}, such as touching nose, combing hair with hand, adjusting eyeglasses, using smartphones, and reading magazines. 

We used ELAN \cite{b14} to annotate the video. The data are annotated into 3 classses, namely eating, drinking, and other. The definitions of eating and drinking gesture are mentioned in Section \uppercase\expandafter{\romannumeral1}. The annotation work was performed by two annotators who followed the same instruction. One is the first author, the other one is independent of this research. The task was divided by sequence. The first author reviewed the annotation from the independent annotator and adjusted it if needed to minimize ambiguity. There are 4,132 eating gestures and 893 drinking gestures in total. The dataset statistics are indicated in Table \ref{eat_drink_data}. Fig. \ref{rd_scene} shows the RD frames of eating and drinking gestures.

\begin{figure*}[t]
      \centering
      \includegraphics[scale=0.65]{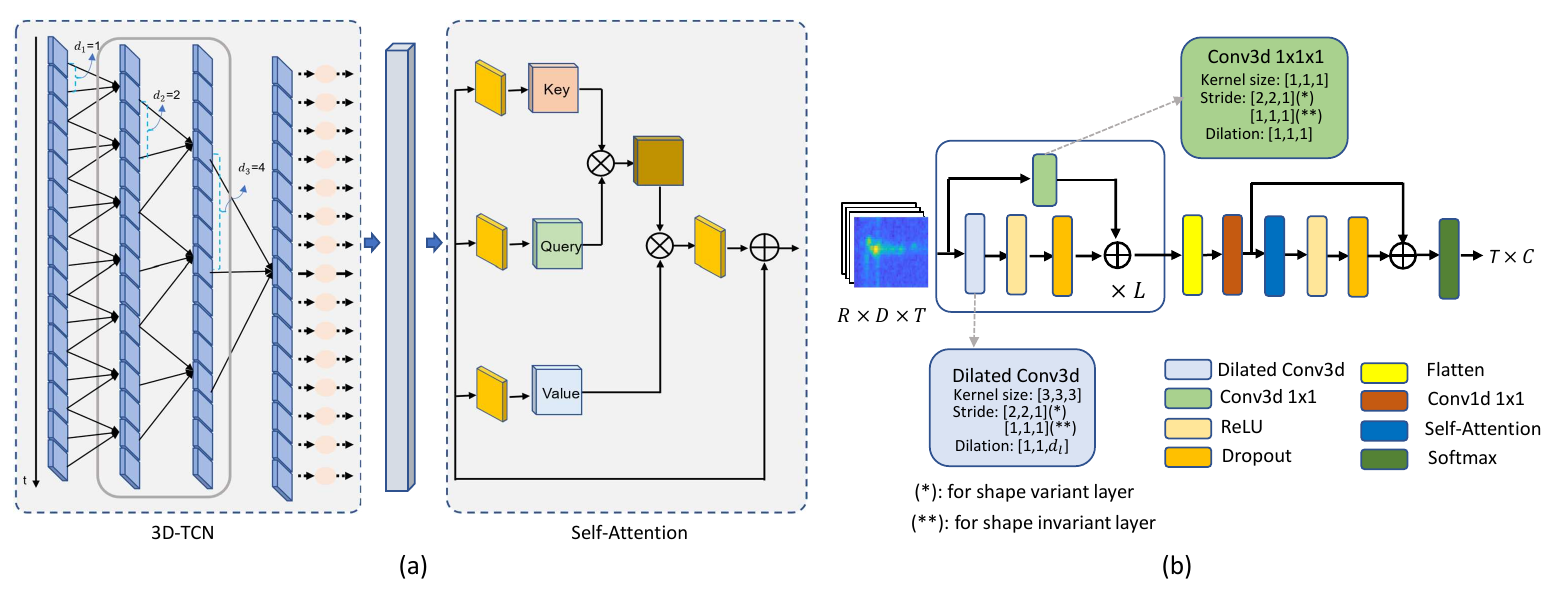}      
      \vspace{-0.2cm}
      \caption{(a) The framework of the 3D-TCN-Att model. From left to right, the RD cube is sent into the input layer (the first layer) of the 3D-TCN module. The extracted features from 3D-TCN are fed into the attention module for prediction refinement. (b) The detailed architecture of the shape-variant (*) and shape-invariant (**) dilated residual layer in 3D-TCN module. The main difference between the two layers is the stride setting. $d_l$ represents the dilated factor, $d_l =2^{l-1}$ $(1\leq{l}\leq{L})$, where $l$ denotes the order of layers, $L$ is the amount of layers. The R, D, T from the input dimension represent the range (R), doppler (D), and time (T). The $C$ in output is the number of classes.}   \vspace{-0.3cm}
      \label{model}
\end{figure*}  

\subsection{Sequence-to-Sequence Neural Network Architecture}

Before feeding the data into the model, prior works \cite{b7,b8,b9} tend to use a sliding window to generate data segments with fixed temporal length. Each window is assigned a label through a majority voting mechanism. Alternatively, they label the window by checking if it contains an entire target activity. We found both approaches have limitations for real-world continuous eating gesture detection. The challenge lies in the variability of eating gesture durations and the fixed length of the sliding window. Firstly, if the duration of the eating gesture is longer than the window length, this gesture may be divided into two or more segments. The segment that contains the minor part of the eating gesture may be labeled as other. Secondly, if a single window encompasses multiple eating gestures with a short duration, it becomes challenging to accurately quantify the total number of eating gestures. To mitigate this issue, we choose the seq2seq architecture instead of the many-to-one architecture. For the seq2seq model, the prediction is generated for each frame rather than for an entire window, referred to as frame-wise prediction. Afterward, a sequence of frame-wise predictions sharing the same value is considered as a segmental interval.

The temporal convolutional network (TCN) using dilated convolution and skip connection was originally introduced by Lea \textit{et al.} \cite{b15} to recognize long-range temporal sequences. To date, it has been utilized to process time-series signals in the healthcare and disease diagnosis domains \cite{b16,b17, b18} due to its outstanding performance \cite{b19}. Meanwhile, the self-attention exhibits superior performance in natural language processing (NLP) \cite{b50} and video-based HAR\cite{b53}. Inspired by its success in \cite{b52,b54}, we integrated the multi-head self-attention into the TCN model, composing the 3D-TCN-Att model to further enhance the temporal sequence processing ability. Noted that existing TCNs \cite{b15,b31} are not suitable in our case, as we employ 3D data. Therefore, the 3D-TCN using Conv3D is developed to process time-series RD Cube. 
\subsubsection{3D-TCN Module}
As illustrated in Fig. \ref{model}, the 3D-TCN module contains a series of $L$ dilated residual layers, where $L$ is the number of layers. Each layer consists of dilated convolutions with ReLU activation, and a residual connection that combines the input of the current layer and the convolution results, as depicted in Fig. \ref{model}(b). The Conv3D kernel with size $3\times3\times3$ convolves input data in range (R), doppler (D), and time (T) dimensions. The number of Conv3D kernels in each layer is 32. The 3D-TCN contains 2 types of dilated layers, the shape-invariant layer and shape-variant layer. The shape-variant layer (Fig. \ref{model}(b)) is used to reduce data shape. Specifically, the size of the range and doppler dimensions are reduced whereas the time dimension remains by setting the stride as $(2,2,1)$. The residual connection has the same stride to keep the same reduction. In contrast, the input and output dimensions of shape-invariant layer remain the same by setting the stride as $(1,1,1)$. The convolutions in both layers are in SAME mode. Finally, the dilation only applies in the time dimension; for the range and doppler dimension, the dilation rate is 1. The dilation factor of the time dimension in Conv3D at each layer is doubled such that $d_l =2^{l-1}$ $(1\leq{l}\leq{L})$. After each dilated layer, a 0.3 dropout is applied.

\begin{table}[t]
\caption{The architecture of 3D-TCN-Att model and the output size of each layer.}
\vspace{-0.2cm}
\label{model_layer}
\begin{center}
\scalebox{0.78}{
\begin{threeparttable} 
\begin{tabular}{llr}
\toprule
 \hline
Layer & Type$^a$ & Output\\
\midrule
Input$^b$&-&$1\times32\times64\times1000$ \\
Layer 1 &Shape variant$^c$ &$32\times32\times32\times1000$\\
Layer 2 &Shape variant &$32\times16\times16\times1000$\\
Layer 3 &Shape variant &$32\times8\times8\times1000$\\
Layer 4 &Shape variant &$32\times4\times4\times1000$\\
Layer 5 &Shape invariant &$32\times4\times4\times1000$\\
Layer 6 &Shape invariant &$32\times4\times4\times1000$\\
Layer 7 &Shape invariant &$32\times4\times4\times1000$\\
Layer 8 &Shape invariant &$32\times4\times4\times1000$\\
Layer 9 &Shape invariant &$32\times4\times4\times1000$\\
Flatten &-& $512\times1000$\\
Conv1D \& Permute &-& $1000\times3$\\
Self-Attention  &-& $1000\times3$\\
ReLU \& Dropout &-& $1000\times3$\\
Softmax&-& $1000\times3$\\
 \hline
\bottomrule
\end{tabular}
    \begin{tablenotes}    
    \footnotesize              
    \item[a] There are two types of dilated residual layer, namely the shape variant layer and the shape invariant layer, as shown in Fig. \ref{model}(b).       
    \item[b] Here we take a 40 s data segment for example, hence the input data size is $1\times32\times64\times1000$. 
    \item[c] The first shape variant layer is different from others; its stride is [1,2,1], the size of range dimension remains the same to achieve the same size between range and doppler dimension. 
    \end{tablenotes}   
\end{threeparttable} 
}
\end{center}
\vspace{-0.3cm}
\end{table}

\subsubsection{Attention Module}
 The attention function can be described as a mapping between the query ($Q=X_{1}W_{q}$), key ($K=X_{1}W_{k}$), and value ($V=X_{1}W_{v}$), where $X_{1}\in\mathbb{R}^{T\times{d_{in}}}$ is the input of attention module (also the output of 3D-TCN module), $W_{q}$, $W_{k}$ and $W_{v}\in\mathbb{R}^{d_{in}\times{d_{model}}}$, $d_{in}$ is the dimension of output sequence from the 3D-TCN model ($d_{in}=3$), and $d_{model}$ represents the dimension of the attention module. The results obtained from the attention can be described as follows:
\begin{equation}Attention (Q,K,V) =\mathrm{Softmax}\left(\frac{QK^\top}{\sqrt{d_{model}}}\right)V\label{eq4}\end{equation}

The aforementioned calculation only involves one single attention graph, whereas multi-head attention utilizes multiple attention graphs to learn attention weights across multiple aspects. The number of heads ($h$) denotes the number of attention graphs.
\begin{equation}
\begin{split}
 MultiHead (Q,K,V) =\mathrm{Concat}(head_{1},\ldots,head_{h})W^{O}\\
 {\rm where}\ head_{i}= Attention(QW_{i}^{Q},KW_{i}^{K},VW_{i}^{V})
\end{split}
\label{eq5}\end{equation}
where $W^{O}\in\mathbb{R}^{h\cdot d_{h}\times{d_{in}}}$, $W_{i}^{Q}$, $W_{i}^{K}$ and $W_{i}^{V}\in\mathbb{R}^{d_{model}\times{d_{h}}}$, and $d_h$ represents the dimension of each single head attention, with $d_h=d_{model}/h$. After applying a residual connection between the input and output of the multi-head attention and a Softmax activation, the final prediction $Y\in\mathbb{R}^{T\times{C}}$ is obtained, where $C$ is the number of classes ($C=d_{in}$).

The loss function of the 3D-TCN-Att model is a combination of the cross-entropy loss as the classification loss, and a truncated mean squared error (T-MSE) as the smoothing loss. Details of the combined loss function can be found in \cite{b34}. According to experiments, we built the 3D-TCN module with 4 shape variant layers, and 5 shape invariant layers. The self-attention module has 4 heads, each with a dimension of 8 ($d_{model}=32$). To train the model, an RMSprop optimizer with a learning rate of 0.0005 is employed. Notably, the positional encoding (PE) is not incorporated in attention module since we did not observe performance improvement after its inclusion. The architecture information is summarized in Table \ref{model_layer}. All the training, and testing experiments were deployed on an Intel 8-core Xeon Gold 6240 CPU@2.6 GHz (Cascadelake) with 20 GB RAM per core, and two pieces of NVIDIA Tesla V100-SXM2-32GB GPU from Vlaams Supercomputer Centrum (VSC) \footnote{See https://www.vscentrum.be/}.

\section{Evaluation and Experiments}
\subsection{Evaluation Scheme}
\subsubsection{Frame-wise Evaluation}
{
To evaluate a seq2seq model, firstly, frame-wise f1-scores for eating and drinking frames are calculated, as our dataset is unbalanced (Table \ref{meal_data}). Secondly, the Cohen Kappa is applied as an overall frame-wise metric. The equations are given below.
\begin{equation} 
{f1\mbox{-}score_{c=\left\{E, D\right\}}=\frac{2tp_c}{2tp_c+fp_c+fn_c}}
\label{eq9}\end{equation}
\vspace{-0.15cm}
\begin{equation} 
{Kappa=\frac{P_o-P_e}{1-P_e}}
\label{eq10}\end{equation}
\vspace{-0.15cm}
\begin{equation} 
{P_o=\sum_{c=N,E,D}\frac{{tp_{c}}}{Total}}
\label{eq11}\end{equation}
\vspace{-0.15cm}
\begin{equation} 
{P_e=\sum_{c=N,E,D}\frac{{tp_c+fp_c}}{Total}\times\frac{{tp_c+fn_c}}{Total}}
\label{eq12}\end{equation}
where $Total$ is the number of data frames, $c$ represents the class, $N, E, D$ represent the class of other, eating, and drinking gesture, respectively. The tp, fp, and fn are frame-wise true positive, false positive, and false negative, respectively. 

However, in the perspective of HAR, frame-wise evaluation is unable to give us the results such as how many eating/drinking gestures are detected. To address this issue, we also utilized segment-wise evaluation.
}

\begin{figure}[t]
      \centering
      \includegraphics[scale=0.50]{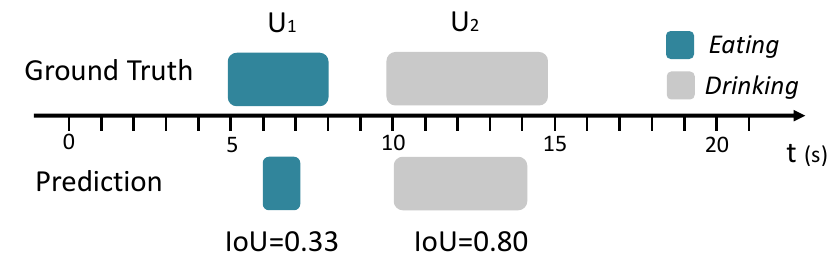}
      \vspace{-0.1cm}
      \caption{The example of IoU calculation.}   
      \label{iou-example}
\end{figure} 

\begin{figure}[t]
      \centering
      \includegraphics[scale=0.40]{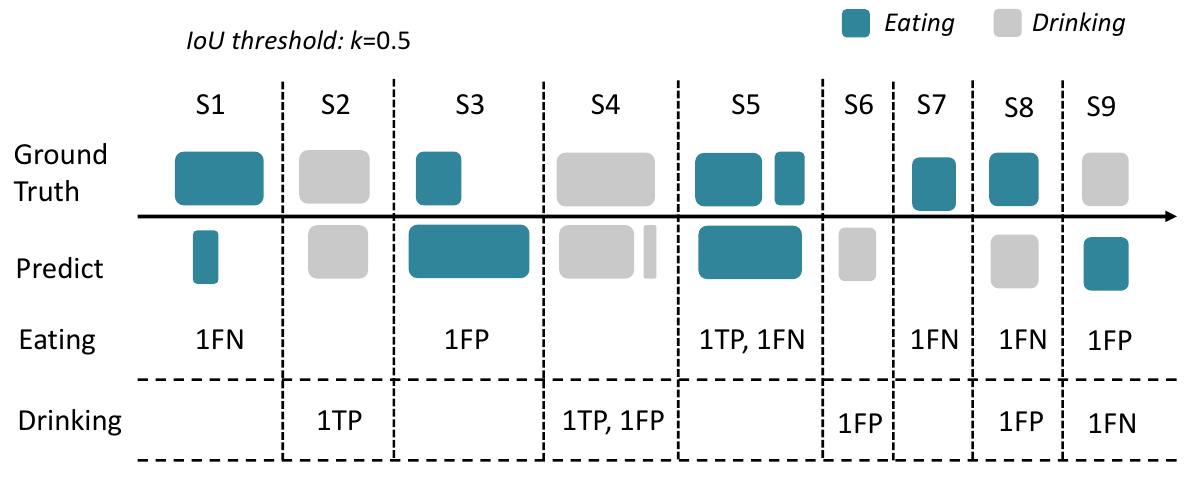}
      \vspace{-0.2cm}
      \caption{Segment-wise evaluation examples when the threshold $k$ is 0.5.}   
      \label{segmental_eva}
\end{figure}
\begin{figure*}[t]
      \centering
      \includegraphics[scale=0.65]{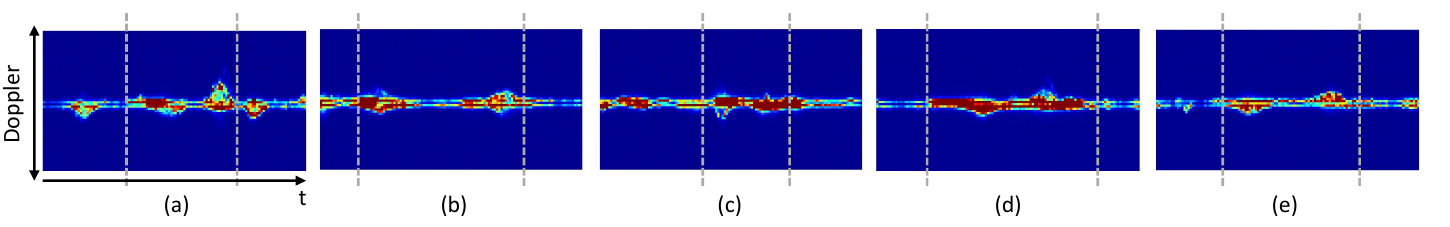}      
      \vspace{-0.2cm}
      \caption{Examples of DT map for different eating and drinking gestures. The duration of each example is 5 s. Signals between grey dashed line represent different eating and drinking gestures: (a) Eating with a pair of fork\&knife; (b) Drinking from a water glass; (c) Eating with a pair of chopsticks; (d) Eating with a spoon; (e) Eating with a hand. }   \vspace{-0.3cm}
      \label{dt_example}
\end{figure*}

\subsubsection{Segment-wise Evaluation}
{The segmental F1-score proposed by \cite{b15, b18} is applied for segment-wise evaluation. The intersection over union (IoU) of each predicted intake gesture is first computed. The IoU is defined as $\frac{A\cap{B}}{A\cup{B}}$, the ratio of overlap between temporal intervals of the ground truth segment (A) and the predicted segment (B). Fig. \ref{iou-example} shows the example of IoU calculation. Once the IoU of each intake gesture is calculated, a threshold $k$ is applied to determine segment-wise True Positive (TP), False Negative (FN), and False Positive (FP). Fig. \ref{segmental_eva} shows evaluation examples. If the IoU of a segment pair is greater than the threshold $k$, it is a TP (Fig. \ref{segmental_eva} S2), otherwise it is an FN segment (Fig. \ref{segmental_eva} S1) or FP segment (Fig. \ref{segmental_eva} S3). The decision tree is as follows:
\begin{equation}Segment=\begin{cases}
TP,&{IoU}\geq{k}\\
FP,&{IoU}<k, length_{gt}<length_p\\
FN,&{IoU}<k, length_{gt}>length_p\\
\end{cases}
\label{eq13}\end{equation}
 where $length_{gt}$ and $length_p$ are the time lengths of the ground truth and predicted eating/drinking gesture, respectively. Additionally, if multiple predicted segments exist inside the range of a given ground truth gesture, only one is classified as a TP, all others are FPs (Fig. \ref{segmental_eva} S4). Conversely, if a predicted segment spans multiple ground truth eating/drinking gestures, only one is recognized as TP and the remainder are FNs (Fig. \ref{segmental_eva} S5). We also defined the corresponding punishment for the misclassification between the eating and drinking gesture (see Fig. \ref{segmental_eva} S8-S9). After the generation of segmental TP, FP, and FN, the segmental F1-score is calculated as:
 \begin{equation}
F1\mbox{-}score_{c=\left\{E, D\right\}}=\frac{2TP_c}{2TP_c+FP_c+FN_c}
\label{eq14}\end{equation}
where $c$ represents the class, $E$, $D$ represents eating and drinking class, respectively.

The segment-wise method provides three main advantages. Firstly, it effectively addressed both over-segmentation errors (Fig. \ref{segmental_eva} S4) and under-segmentation errors (Fig. \ref{segmental_eva} S5); secondly, it accommodates minor temporal discrepancies between the ground truth and prediction, which can occur due to annotation variability. Thirdly, it provides clear information including how many gestures are detected. Three thresholds $k$ are selected as 0.1, 0.25 and 0.5 according to \cite{b15}. 
}

\subsection{Models For Benchmark}
Different deep learning models are selected as benchmarks. Specifically, the CNN-(Bi)LSTM \cite{b20,b21}, pretrained 3D-CNNs, and transformers. It should be noted that the output modes of all models are configured as sequence output.
\subsubsection{CNN-(Bi)LSTM}
The CNN-(Bi)LSTM model consists of three subnetworks: three CNN layers, one LSTM layer, and two fully connected layers (FCN). The CNN part contains three layers with 16, 32, and 32 kernels with size $3\times{3}$, respectively. The activation ReLU is used for each layer. Each CNN layer is followed by a maxpooling layer with $2\times{2}$ pooling size. The output of the CNN part is flattened and fed into the LSTM part. The LSTM layer contains 128 hidden units followed by a 0.3 dropout. The FCN contains two dense layers. The first layer has 128 neurons with ReLU activation, and the second layer has 3 neurons with Softmax activation. The Timedistributed layer is used to wrap the model. Based on experiments, the window length of CNN-LSTM and CNN-BiLSTM is 10 s and 20 s, respectively.

\subsubsection{Pretrained 3D-CNNs}
In video-based HAR domain, numerous 3D-CNN-based models have been proposed and pretrained on large scale RGB-based datasets \cite{b39}. We conducted experiments using four widely used pretrained 3D-CNN models: 3DResNet-18 \cite{b35}, 3DResNet-50 \cite{b36}, X3D \cite{b37}, R(2+1)D \cite{b38}. These models were modified to fit our data processing pipeline. The 3DResNet-18 and R(2+1)D models were built based on Torchvision package, while the 3DResNet-50 and X3D models were from FAIR PyTorchVideo package. All models were pretrained on the Kinetics dataset \cite{b49}. In our experiments, we modified the output layers for frame-wise prediction and adjusted the number of predicted classes to 3. The window length of these models is 10 s. All models were trained for 100 epochs with the RMSprop optimizer, a learning rate of 0.0005, and a batch size of 64.

\subsubsection{Transformers}
Two transformer models were employed for our experiments: the vanilla Transformer \cite{b50}, and the ViViT model \cite{b48}. Based on experiments, both were trained from scratch. The vanilla Transformer has the same encoder architecture as introduced in \cite{b50}; however, as it can only process 1D data, a 3-layer CNN module is used to extract spatial features from 3D radar data and flatten it into 1D intermediate data. The added 3-layer CNN is identical to the CNN component used in the CNN-LSTM model. Hence, the CNN-vanilla Transformer is called CNN-VT in this study. Meanwhile, based on our experiments, the number of encoder layer in CNN-VT model is 1, with 8 heads, each with the dimension of 64. For the ViViT model, the patch size for each input frame is 16, the number of transformer layer is 6, and each layer has 8 heads with a head dimension of 64. The window length, number of epochs, and learning rate are consistent with those of pretrained 3D-CNNs.

\subsection{Radar Data Representations}
For radar-based HAR systems, another common radar data representation is the DT map. Compared to RD Cube that contains range, velocity, and time dimensions,  the DT map only includes information from time and velocity dimensions. For the eating activity, apart from velocity, the range information of the arm and the head (or torso) is also an essential feature. Thus, this research focuses on using RD Cube for eating gesture detection. To validate our hypothesis, we also test the DT map as input, with the dimension $N_f\times{N_{d1}}$. Fig. \ref{dt_example} shows the examples of DT map. Due to the dimension difference between the DT map (2D) and the RD Cube (3D), proposed models used for the RD Cube (3D-TCN, CNN-(Bi)LSTM) are modified as 2D types to fit the DT map.

\subsection{Validation Strategy}
To evaluate the model performance on unseen data, the dataset is divided into 7 folds. For each fold, the number of meals in the train/valid/test set is 50/10/10, respectively. It should be noted that the split of dataset in this validation method is meal level, there is no information overlap between the train, valid, and test set. (The meal\_id lists for 7-fold experiments are available along with the dataset.)

\begin{table}[t]
\caption{Overall frame-wise results for 7-fold cross-validation}
\vspace{-0.2cm}
\label{frame_results}
\begin{center}
\scalebox{0.75}{
\begin{tabular}{c|l|c|c|c}
\toprule
\hline
\multirow{2}*{Data}&\multirow{2}*{Model}&\multicolumn{2}{|c|}{f1-score}&\multirow{2}*{Kappa}\\
~&~&E&D &~\\
\midrule
\multirow{3}*{DT map }&CNN-LSTM&0.519  &0.523 & 0.450\\
~&CNN-BiLSTM&0.570  &0.636 & 0.523\\
(2D)&2D-TCN&0.742 &0.805  & 0.715 \\
\midrule
\multirow{9}*{RD Cube }&CNN-LSTM&0.654  &0.697 & 0.605 \\
~&CNN-BiLSTM&0.713 &0.763 & 0.672  \\
~&3DResNet-18 &0.699 &0.720 & 0.653  \\
~&3DResNet-50 &0.588 &0.575 & 0.520  \\
~&X3D  &0.716 &0.732 & 0.667  \\
(3D)&R(2+1)D &0.696 &0.682 & 0.636  \\
~&CNN-VT&0.654 &0.666  & 0.589 \\
~&ViViT&0.577 &0.474 & 0.449  \\
~&3D-TCN&0.772 &\textbf{0.819}  & 0.736 \\
~&3D-TCN-Att&\textbf{0.776} &0.817  & \textbf{0.745}  \\
\hline
\bottomrule
\end{tabular}}
\end{center}
\vspace{-0.2cm}
\end{table}

\begin{table*}[t]
\caption{Overall segment-wise results for 7-fold cross-validation}
\vspace{-0.2cm}
\label{seg_results}

\begin{center}
\scalebox{0.82}{
\begin{threeparttable} 
\begin{tabular}{c|l|ccc|ccc|l|l}
\toprule
\hline
\multirow{2}*{Data}&\multirow{2}*{Model}&\multicolumn{3}{|c|}{Eating Segmental F1-score}&\multicolumn{3}{|c|}{Drinking Segmental F1-score}&\multirow{2}*{\#Params}&\multirow{2}*{Latency}\\
~&~&{$k=0.1$}&{$k=0.25$}&{$k=0.5$}&{$k=0.1$}&{$k=0.25$}&{$k=0.5$}&~&~\\
\midrule
\multirow{3}*{DT map }&CNN-LSTM&0.735&	0.715&	0.609&	0.745&	0.696&	0.578 &0.228 M$^a$&\textbf{0.021 s}\\
~&CNN-BiLSTM&0.763&0.749 & 0.669&0.813  &0.782 & 0.694&0.442 M&0.031 s\\
(2D)&2D-TCN&0.890&0.885 & 0.855&0.887&	0.880&	0.858&\textbf{0.115 M}&0.085 s\\
\midrule
\multirow{9}*{RD Cube }&CNN-LSTM&0.843&0.837 & 0.766&0.816  &0.795 & 0.740&0.294 M& 0.217 s\\
~&CNN-BiLSTM&0.884&	0.877&	0.828&	0.866&	0.852&	0.812&0.573 M& 0.233 s\\
~&3DResNet-18 &0.873&	0.865&	0.807&	0.832&	0.805&	0.732&33.353 M& 20.582 s\\
~&3DResNet-50 &0.798&	0.789&	0.729&	0.642&	0.614&	0.554&31.634 M&5.871 s\\
~&X3D &0.890&	0.883&	0.834&	0.825&	0.819&	0.785&2.980 M& 1.870 s\\
(3D)&R(2+1)D  &0.844&	0.840&	0.792&	0.748&	0.725&	0.655&31.297 M& 15.398 s\\
~&CNN-VT&0.846&	0.833&	0.768&	0.760&	0.749& 0.699&1.667 M & 0.485 s\\
~&ViViT &0.749&	0.728&	0.645&	0.603&	0.570&	0.501&7.235 M & 0.855 s\\
~&3D-TCN&0.921&	0.918&	0.890&	0.888 &0.883 & 0.864&0.300 M & 0.991 s\\  
~&3D-TCN-Att&0.924&	0.922&	\textbf{0.896}&	0.893 &0.889 & \textbf{0.868}&0.302 M & 1.062 s\\ 
\hline
\bottomrule    
\end{tabular}
    \begin{tablenotes}    
    \footnotesize              
    \item[a] The M represents million. So 0.228 M represents 0.228 million parameters.         %       
    \end{tablenotes}   
\end{threeparttable} 
}
\end{center}
\vspace{-0.6cm}
\end{table*}

\begin{table}[t]
\caption{Average Results Across 7-fold cross-validation ($k=0.5$)}
\vspace{-0.2cm}
\label{average_results}
\begin{center}
\scalebox{0.78}{
\begin{threeparttable} 
\begin{tabular}{c|l|c|c|c|c}
\toprule
 \hline
\multirow{2}*{Data}&\multirow{2}*{Model}&\multicolumn{2}{|c|}{F1-score (mean$\pm$std)}&\multicolumn{2}{|c}{$p^a$}\\
~&~&E&D &E&D\\
\midrule
\multirow{3}*{DT map }&CNN-LSTM&0.594$\pm$0.094  &0.575$\pm$0.083 & ***&***\\
~&CNN-BiLSTM&0.670$\pm$0.081  &0.693$\pm$0.037 & ***&***\\
(2D)&2D-TCN&0.853$\pm$0.031  &0.856$\pm$0.078 & **&0.23\\
\midrule
\multirow{9}*{RD Cube }&CNN-LSTM&0.768$\pm$0.031  &0.731$\pm$0.080 & ***&***\\
~&CNN-BiLSTM&0.826$\pm$0.024  &0.806$\pm$0.089 & **&**\\
~&3DResNet-18 &0.807$\pm$0.024  &0.727$\pm$0.070 & ***&***\\
~&3DResNet-50 &0.722$\pm$0.081  &0.583$\pm$0.111 & ***&***\\
~&X3D &0.830$\pm$0.041 &0.777$\pm$0.099& **&***\\
(3D)&R(2+1)D &0.779$\pm$0.077 &0.663$\pm$0.156 & ***&***\\
~&CNN-VT&0.768$\pm$0.033  & 0.699$\pm$0.060 & ***&***\\
~&ViViT&0.645$\pm$0.040  & 0.498$\pm$0.051 & ***&***\\
~&3D-TCN&0.888$\pm$0.037  & 0.857$\pm$0.061 & *&0.08\\
~&3D-TCN-Att&\textbf{0.895$\pm$0.035}  & \textbf{0.861$\pm$0.054} & -&-\\
 \hline
\bottomrule
\end{tabular}
    \begin{tablenotes}    
    \footnotesize              
    \item[a] The p values are obtained by applying paired t-test (7-fold experiment) between 3D-TCN-Att and other models for eating and drinking detection performance. $p<0.05$: *, $p<0.01$: **, $p<0.001$: ***.         %       
    \end{tablenotes}   
\end{threeparttable} 
}
\end{center}
\vspace{-0.3cm}
\end{table}

\begin{table}[t]
\caption{Overall misclassification results}
\vspace{-0.2cm}
\label{misclass_results}
\begin{center}
\scalebox{0.78}{
\begin{threeparttable} 
\begin{tabular}{c|l|l|l}
\toprule
\hline
Data&Model&\#(E$\rightarrow$D)$^a$&\#(D$\rightarrow$E)\\
\midrule
\multirow{2}*{DT map}&CNN-LSTM& 70 (1.69\%)&32 (3.58\%)\\
~&CNN-BiLSTM&39 (0.94\%)&24 (2.69\%)\\
(2D)&2D-TCN&23 (0.56\%)&\textbf{10 (1.12\%)}\\
\midrule
\multirow{9}*{RD Cube}&CNN-LSTM&25 (0.61\%)&25 (2.80\%)\\
%\midrule
~&CNN-BiLSTM&22 (0.53\%)&20 (2.24\%)\\
~&3DResNet-18 &66 (1.60\%)&28 (3.14\%)\\
~&3DResNet-50 &227 (5.49\%)&104 (11.65\%)\\
~&X3D &61 (1.48\%)&26 (2.91\%)\\
(3D)&R(2+1)D  &90 (2.18\%)&63 (7.05\%)\\
~&CNN-VT&61 (1.48\%)&26 (2.91\%)\\
~&ViViT&94 (2.27\%)&107 (11.98\%)\\
~&3D-TCN&\textbf{19 (0.46\%)}&24 (2.69\%)\\
~&3D-TCN-Att&23 (0.56\%)&24 (2.69\%)\\
\hline
\bottomrule
\end{tabular}
    \begin{tablenotes}    
    \footnotesize              
    \item[a] \#(E$\rightarrow$D) denotes the number of ground truth eating gesture predicted as drinking gesture (Fig. \ref{segmental_eva} S8). \#(D$\rightarrow$E) represents the reverse case (Fig. \ref{segmental_eva} S9).  
    \end{tablenotes}   
\end{threeparttable} 
}
\end{center}
\vspace{-0.6cm}
\end{table}

\begin{table}[t]
\caption{Performance of eating gesture detection for four eating styles using 3D-TCN-Att ($k=0.5$)}
\vspace{-0.2cm}
\label{type_results}
\begin{center}
\scalebox{0.8}{
\begin{tabular}{ccccc}
\toprule
 \hline
& Fork \& knife&Chopsticks&Spoon&Hand\\
\midrule
F1-score&0.897&0.920&0.901&0.813\\
 \hline
\bottomrule
\end{tabular}}
\end{center}
\end{table}

\section{Results}
Table \ref{frame_results} presents the frame-wise results on DT map and RD Cube. Both the f1-scores and kappa scores from three models (CNN-LSTM, CNN-BiLSTM, and 2D/3D-TCN) in RD Cube were higher than those of in DT map. Using DT map data, the 2D-TCN obtained the highest performance with a kappa of 0.715, an eating f1-score of 0.742, and a drinking f1-score of 0.805. For RD Cube, the 3D-TCN-Att obtained the highest kappa of 0.745 and eating f1-score of 0.776. The 3D-TCN yielded the best drinking f1-score of 0.819.

Table \ref{seg_results} shows the overall segment-wise results. In segment-wise evaluation, the RD Cube still obtained better performance on all three models (CNN-LSTM, CNN-BiLSTM, and 2D/3D-TCN). For $k=0.1$, the 3D-TCN-Att model yielded the highest F1-score of 0.924 for eating and 0.893 for drinking, compared to the X3D (E: 0.890; D: 0.825) and the CNN-BiLSTM (E: 0.884; D: 0.866). When the threshold was raised to 0.5, the performance of all models decreased. However, 3D-TCN-Att still obtained the best results (E: 0.896; D: 0.868) with the least reduction (E: 2.8\%, D: 2.5\%), compared to X3D (E: 0.890$\rightarrow$0.834; D: 0.825$\rightarrow$0.785).

\begin{table*}[t]
\caption{Comparing the literature on in-meal eating gesture detection}
\vspace{-0.2cm}
\label{compare_litera}
\begin{center}
\scalebox{0.85}{
    \begin{tabular}{l|l|l|l|l|l|l}
    \toprule
    \hline
        Modality & Study/Dataset & \#Subjects& \#Meal& Model  & Validation &Performance \\ 
\midrule
        \multirow{3}*{IMU }&  FIC \cite{b5} & 12 & 21 & CNN-LSTM & LOSO& Eating F1: 0.923 \\ 
        ~&  OREBA-IMU \cite{b4}& 100 & 100 & ResNet-10 CNN-LSTM & train:valid:test=61:20:19&Eating F1: 0.837 Drinking F1: 0.770\\ 
        ~&  Clemson \cite{b43} & 271 & 271 & Non-machine learning  & - & Eating F1: 0.814\\ 
        \midrule
       \multirow{2}*{Video}& Hossain \textit{et al.} \cite{b44} & 28 & 84 & Faster R-CNN+AlexNet  & 4-fold & Eating accuracy: 0.854 \\ 
        ~&  OREBA-Video \cite{b4} & 100 & 100  & ResNet-50 CNN-LSTM & train:valid:test=61:20:19& Eating F1: 0.869 Drinking F1: 0.761\\ 
        \midrule
        \multirow{2}*{Acoustic}& Sazonov \textit{et al.} \cite{b45} & 20 & 70  & SVM & 3-fold & Average weighted accuracy: 0.847  \\ 
        ~ &  K. Lee \cite{b46} & 10 & 40 & ANN  & train:test=7:3& Chewing accuracy: 0.914 Swallowing accuracy: 0.784 \\ 
        \midrule
        Radar&  Proposed & 70 & 70 & 3D-TCN-Att  & 7-fold & Eating F1: 0.896 Drinking F1: 0.868 ($k=0.5$)\\ 
\hline
\bottomrule
\end{tabular}}
\end{center}
\vspace{-0.3cm}
\end{table*}

In addition to segmental F1-scores, the number of parameters for each model and their inference time (the latency) were also indicated in Table \ref{seg_results}. Regarding DT map data, the amount of parameters in 2D-TCN was the least (0.115M). For RD Cube data, the number of parameters in all pretrained 3D-CNN models was significantly larger than CNN-LSTM (0.294 M) and 3D-TCN-Att models (0.302 M). To assess the model’s real-time ability after deployment, we conducted tests to measure the inference time for processing 40 s data. Models were deployed on a laptop with Intel Core i7 10750 CPU @2.6 GB, 6 cores (without GPU). Results in Table \ref{seg_results} represent the average inference time across 100 test runs. The CNN-LSTM had the lowest latency compared to other models (0.217 s). The 3D-TCN-Att required 1.062 s to generate predictions.

Table \ref{average_results} shows the average segmental F1-score ($k=0.5$) across 7-fold cross-validation experiment. To further compare the performance difference between 3D-TCN-Att and other models, paired t-test was applied. For eating gesture detection, the 3D-TCN-Att obtained the highest mean F1-score (0.895$\pm$0.035, $p<0.05$). For drinking gesture detection, the model still yielded the highest mean F1-score with statistical significance (0.861$\pm$0.054, $p<0.05$) for all models except 2D-TCN ($p=0.23$) and 3D-TCN ($p=0.08$).

Table \ref{misclass_results} presents the number of misclassifications between eating and drinking gestures. The 3D-TCN-Att model wrongly predicted 23 ground truth eating gestures (0.56\% of total ground truth eating gestures) as drinking gestures (E$\rightarrow$D, see Fig. \ref{segmental_eva} S8), which was slightly higher than that of the 3D-TCN model (0.46\%). For the case of misclassifying drinking as eating (D$\rightarrow$E, see Fig. \ref{segmental_eva} S9), 2D-TCN had the least number.

Four different eating styles were included in our dataset (Table \ref{eat_data}). To investigate if the eating style has any impact on the performance, the segment-wise F1-score for each eating style on 3D-TCN-Att with RD Cube data was calculated, as shown in Table \ref{type_results}. The 3D-TCN-Att obtained the highest F1-score for eating with chopsticks (0.920). The detection of eating with hand had the lowest performance (0.813) compared to the other three eating gestures. It should be clarified that this analysis did not involve eating style classification. The results were obtained from the 7-fold validation. The entire dataset (70 meals) was categorized into four groups based on eating styles to facilitate this analysis.

\begin{figure}[t]
      \centering
      \includegraphics[scale=0.4]{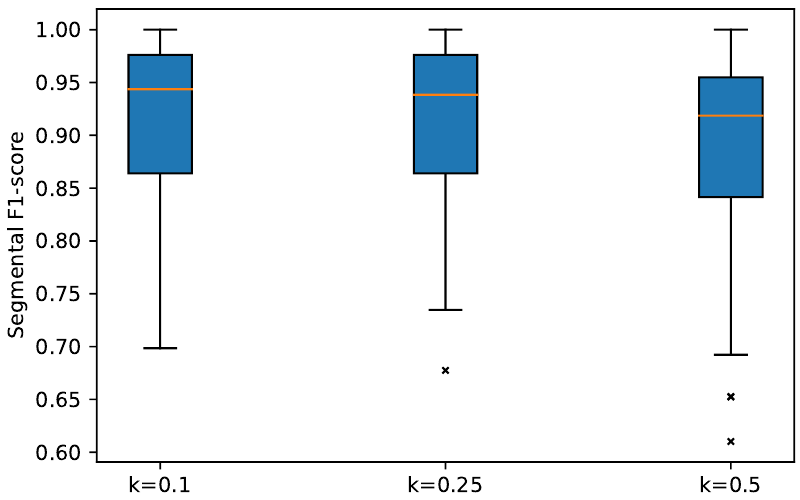}      
      \vspace{-0.1cm}
      \caption{The box plot of the results from 70 meals. The outlier with the lowest F1-score for $k=0.5$ is the meal eaten with hand. }   \vspace{-0.2cm}
      \label{results_line}
\end{figure}

Fig. \ref{results_line} denotes the segmental F1-score distribution for 70 meal sessions. All the meals had F1-scores higher than 0.7 for $k=0.1$. When $k$ was increased to 0.5, one meal session's F1-score was equal to or lower than 0.6 (the lowest outlier), which was eaten by hand.

\section{Discussion}
The proposed radar-based food intake monitoring system obtained promising results. There are several elements that distinguish our work from existing approaches \cite{b10,b33}. Firstly, we developed an end-to-end framework for eating behavior detection in continuous meal sessions. In contrast to previous radar-based works, continuous meal sessions involve lots of other non-feeding activities, which are seamlessly performed together with intake gestures during meal sessions. These activities are continuous, with varying durations and natural transitions between activities \cite{b30}, making the eating gesture detection in meal session more challenging, but more meaningful in terms of real-life application \cite{b42}. Secondly, we evaluated our model on a significantly larger dataset with more participants (70 vs. 6). Thirdly, we detected intake gestures from a dataset that includes various other classes of activities, whereas the work in \cite{b10} focused on the classification between 5 eating styles (drinking and other classes were not reported in their data.). For the study in \cite{b33}, the duration ratio of eating and non-eating activity was 1.2 : 1 (eating : non-eating). However, in real meal sessions, intake gestures occur sparsely, resulting in an imbalanced dataset (ratio of duration for drinking : eating : others = 1 : 2.59 : 11.21 in our case).

When comparing our study to existing food intake monitoring studies using different sensors but with a similar experiment setup, as presented in Table \ref{compare_litera}, the size of our dataset is comparable to others. Furthermore, the radar performance aligns with existing approaches (IMU-, video-, and acoustic-based sensor) in similar settings, which provides evidence that the radar is a promising modality for further investigation.

Currently, the conventional method in HAR is using a sliding window and there is only one label per window. In real-life scenarios, the duration of intake gestures and the time distance between two intake gestures vary greatly (Table \ref{eat_drink_data}). It is difficult to annotate the window as eating, drinking or other, because it is possible that one window can contain multiple short eating gestures, or only a part of one eating gesture. Hence we applied the more desirable seq2seq architecture to obtain frame-wise predictions. Frame-wise labels are used in the training process, the seq2seq architecture can also enable the ability of segmentation for detected gestures. 

Several models were implemented to evaluate the performance. Results show that the performance of pretrained 3D-CNNs is lower than that of 3D-TCN-Att. One potential reason is the substantial dissimilarity between the pretrained video dataset and our radar dataset. Although both are 3D data, the information provided by video and RD cube is fundamentally different. As for transformers, they require large-scale datasets and significant computational resources due to their lack of inductive biases. The success of transformers in video domain relies on extensive training on large-scale datasets. However, radar-based HAR is still at an early stage compared to video-based study, resulting in a scarcity of large-scale radar datasets. Due to the limited data size, we did not take full advantage of pretrained 3D-CNNs and transformers, which are worth for further investigation in radar domain.

The CNN-(Bi)LSTM models exhibited significantly lower latency compared to others. This is primarily because CNN-RNNs utilized Conv2D, while others relied on Conv3D, which required more computational resources. Although 3D-TCN-Att had a higher latency (1.062 s) compared to CNN-RNNs, it was still much shorter than the waiting time for integrating received data (40 s), which implies that mainstream computers can support the proposed method in real-time scenarios. The Conv3D enabled 3D-TCN-Att to extract and process spatial-temporal features simultaneously, resulting in better performance at the cost of increased computation. Using specially designed 2D CNN to learn spatial-temporal representations can be a potential solution to mitigate the cost \cite{b47}.

The performance of using RD Cube data was higher than that of using DT map data. This result is consistent with our hypothesis in Section \uppercase\expandafter{\romannumeral3}-C, which suggests that the range information between head and hand is an important feature for radar-based food intake detection.

\begin{figure}[t]
      \centering
      \includegraphics[scale=0.67]{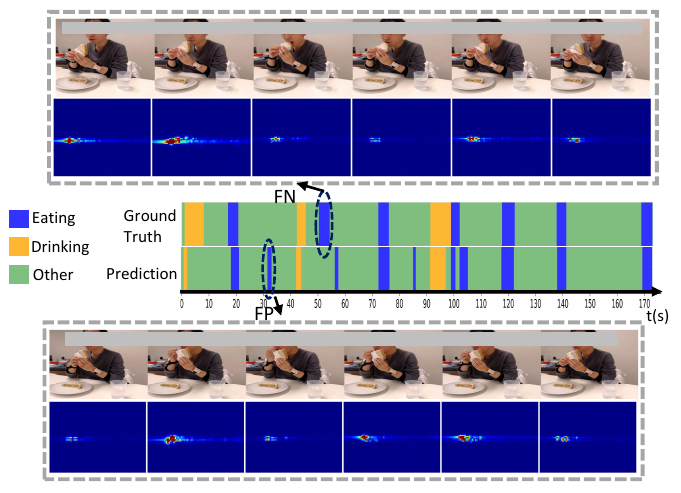}
      \vspace{-0.1cm}
      \caption{Failure examples for the lowest outlier meal in Fig. \ref{results_line}. In the FP example, the participant used his right hand to touch nose while using the left hand to hold the sandwich, which was similar to delivering food to mouth by hand. In the FN example, the participant rotated his wrists to eat food without obvious arm and body movement. }   \vspace{-0.3cm}
      \label{bad_example}
\end{figure}

The performance of drinking gesture detection was lower than that of eating gesture detection. One reason is that the amount of drinking gestures was much less than that of eating (893 vs. 4132). Meanwhile, the detection performance for the gesture of eating with hand was the lowest. One potential reason is that there is more variety in eating with hands. Moreover, by investigating misclassification cases, we found that 3.43\% of the gestures belonging to eating with hand were predicted as drinking, which was higher than the other three eating gestures (0.37\% for fork \& knife, 0.14\% for chopsticks, 0.30\% for spoon). This observation suggests that the gesture of eating with hand is similar to the drinking gesture.

We investigated the meal that obtains the lowest F1-score (the lowest outlier in Fig. \ref{results_line}) by checking the video. Two typical wrong cases were selected, as shown in Fig. \ref{bad_example}. For the FP example, the participant used one hand to touch his nose, while using the other hand to hold the sandwich close to his mouth, which was similar to eating sandwich with hand. For the FN example, the participant mainly used his wrists to rotate the sandwich to his mouth, no obvious hand/arm movement.

\begin{figure}[t]
      \centering
      \includegraphics[scale=0.45]{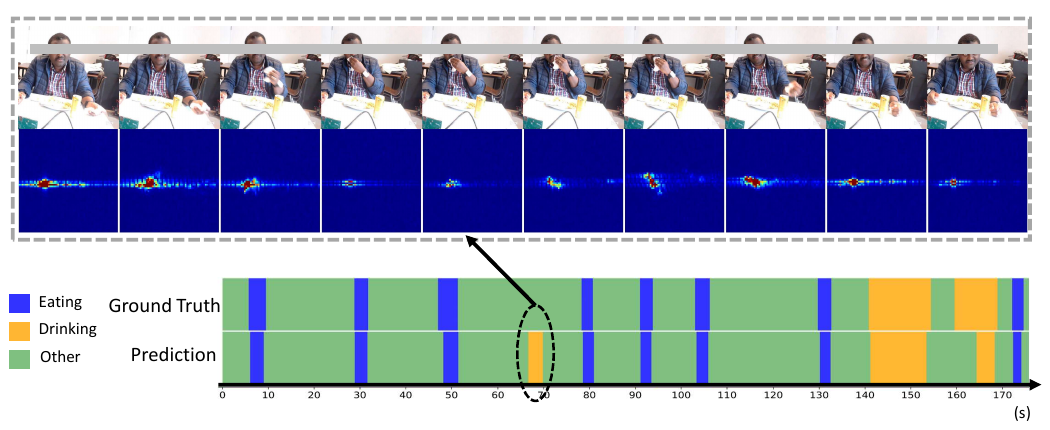}
      \vspace{-0.2cm}
      \caption{An example of failure case. The participant was wiping mouth with tissues. The model predicted it as a drinking gesture. }   \vspace{-0.2cm}
      \label{noise_example}
\end{figure}

Several sensors have been proposed for food intake monitoring, each with its own advantages and limitations. Despite the benefits of radar discussed in Section \uppercase\expandafter{\romannumeral1}, there are still several technical challenges inherent to radar sensor. Firstly, radar data requires complex signal processing procedures compared to IMU and video data. Secondly, there is less context information in radar compared to video. Thirdly, for radar sensor, recognizing eating gestures from multiple people is more challenging if people sit closely and eat at same time. Using an FMCW radar for food intake monitoring has yet to be explored broadly. We propose this method as a good supplement to smartwatch-based and video-based approaches. One application scenario is in-door activity recognition, since video monitoring at home has higher privacy concern and people could not wear a smartwatch all the time. Apart from food intake monitoring solely, this method can also be explored to assess upper limb tremors of the elderly continuously and unobtrusively in real-life meal sessions \cite{b11}.

We collected the dataset at the university, not in a home-like environment. However, since we applied static object removal, the stationary layout information was removed in this step. The results on the data collected in seven different rooms with different layouts indicated that the proposed method can detect eating gestures in different environments.

However, there are still few limitations in this research. Firstly, the data were collected in static environments, which means there were no other moving objects/subjects in the ROI of radar during data collection. The performance in environments containing other moving objects and persons still needs further exploration. A typical radar noise for indoor scenarios caused by multipath problem \cite{b51} should also be considered in further study. Secondly, we observed that the model occasionally misclassifies certain deceptive movements as intake gestures. For instance, actions like wiping the mouth with a napkin, which was similar to drinking gesture from radar signal, as illustrated in Fig. \ref{noise_example}. Thirdly, this study mainly concentrated on the single-subject scenario. The multiple-subject eating scenario should be explored as future work. One potential solution is to incorporate an additional feature dimension, namely the angle information provided by the multiple-input-multiple-output (MIMO) radar. By integrating this information, the Range-Doppler-Angle-Time 4D data can be used to recognize activities from multiple subjects.
\section{Conclusion}
In this research, we proposed Eat-Radar, a continuous food intake detection system, using an FMCW radar sensor and a seq2seq 3D-TCN-Att model. The detection and segmentation of eating and drinking gestures were realized by processing RD Cube from the radar. The effectiveness of this approach was validated on a collected dataset containing 70 meal sessions. The results from the proposed approach and other existing methods explored in our work demonstrated the feasibility of using FMCW radar for food intake monitoring. The proposed end-to-end framework including the seq2seq deep learning model and the segment-wise evaluation method has the potential to be applied in other radar-based HAR research. 
\section*{Acknowledgment}
The authors would like to thank the participants who participated in the experiments for their efforts and time.

\end{document}